%% file: 0500.tex
\documentclass[runningheads]{llncs}
\makeatletter
\newcommand{\@chapapp}{\relax}%
\makeatother

\usepackage{graphicx}
\usepackage{amsmath,amssymb} 
\usepackage{upgreek}
\usepackage{color}
\usepackage{appendix}
\usepackage{cancel}

\usepackage{bm}	
\usepackage{bbm}
\usepackage{epsfig}
\usepackage{siunitx}

\usepackage{algorithm}			
\usepackage{algorithmicx}			

\usepackage[width=122mm,left=12mm,paperwidth=146mm,height=193mm,top=12mm,paperheight=217mm]{geometry}

\usepackage[caption=false]{subfig}
\usepackage{sidecap}

\usepackage{cleveref}

\input{commands.tex}

\begin{document}
\pagestyle{headings}
\mainmatter

\title{Branching Gaussian Processes with Applications to Spatiotemporal Reconstruction of 3D Trees} 

\titlerunning{Branching Gaussian Processes for 3D Tree Reconstruction}

\authorrunning{K. Simek, R. Palanivelu, K. Barnard}

\author {Kyle Simek$^{*}$, Ravishankar Palanivelu$^{\dagger}$, Kobus Barnard$^{\ddagger}$}
\institute{
    $^{*}$\normalsize Matterport, Inc.,\\
    $^{\dagger}$\normalsize School of Plant Sciences, Univ. of Arizona \\
    $^{\ddagger}$\normalsize Computer Science, Univ. of Arizona \\
    \quad\email{kylesimek@gmail.com}\quad
    \quad\email{rpalaniv@email.arizona.edu}\quad
    \quad\email{kobus@cs.arizona.edu}\\
}
\maketitle

\input{abstract}

\input{intro}
\input{related_work}

\input{bgp}

\input{modeling}
\input{inference}
\input{experiments}

\input{conclusion}

\bibliographystyle{splncs03}
\bibliography{kyle_refs,supplemental_refs}

\end{document}

%% file: commands.tex
\def\iid{{i.i.d}\OneDot}
\def\ie{{i.e}\OneDot}
\def\eg{{e.g}\OneDot}

\def\etal{{et al}\OneDot}
\def\iff{if and only if }

\newcommand{\comment}[1]{}

\newcommand{\squishlist}{
   \begin{list}{$\bullet$}
    { \setlength{\itemsep}{0pt}      \setlength{\parsep}{3pt}
      \setlength{\topsep}{3pt}       \setlength{\partopsep}{0pt}
      \setlength{\labelwidth}{1em}
      \setlength{\labelsep}{0.5em} } }

\newcommand{\squishlisttwo}{
   \begin{list}{$\bullet$}
    { \setlength{\itemsep}{0pt}    \setlength{\parsep}{0pt}
      \setlength{\topsep}{0pt}     \setlength{\partopsep}{0pt}
      \setlength{\labelwidth}{1.5em}
      \setlength{\labelsep}{0.5em} } }

\newcommand{\squishlistthree}{
   \begin{list}{$\bullet$}
    { \setlength{\itemsep}{0.01pt}    \setlength{\parsep}{0.01pt}
      \setlength{\topsep}{0pt}     \setlength{\partopsep}{0pt}
      \setlength{\labelwidth}{0.5em}
      \setlength{\labelsep}{0.5em} } }

\newcommand{\squishend}{
    \end{list}  }

\renewcommand*{\th}{\textsuperscript{th }}

\newcommand{\abs}[1]{{\lvert #1 \rvert}}
\newcommand{\Exp}[1]{{\operatorname{exp}\left( #1 \right)}}

\newcommand{\R}{\mathbb{R}}

\newcommand{\nofrac}[2]{#1 / #2}
\newcommand{\dddK}{\dddot{K}}

\newcommand{\N}{\mathcal{N}}
\newcommand{\GP}{\mathcal{GP}}

\newcommand{\distras}{\sim}

\newcommand{\bz}{\bm{z}}
\newcommand{\bZ}{\bm{Z}}

\newcommand{\xc}{c}
\newcommand{\xt}{t}
\newcommand{\T}{\uptau}
\def\C{\mathcal{C}}
\def\D{\bm{D}}
\newcommand{\parent}{\rho}
\newcommand{\parentC}{\rho_\C}
\newcommand{\parentT}{\rho_\T}
\newcommand{\I}{\mathcal{I}} 

\newcommand{\bgp}{\mathrm{BGP}}
\newcommand{\tbgp}{\mathrm{tBGP}}

\def\mm{\si{\milli\meter\,}}
\def\cm{\si{\centi\meter\,}}
\def\Arabidopsis{\emph{Arabidopsis }}

\def\KL{Kullback-Leibler }

\def\polybezier{poly-B\'ezier }

%% file: abstract.tex
\begin{abstract}
We propose a robust method for estimating dynamic 3D curvilinear branching structure from monocular images.
While 3D reconstruction from images has been widely studied, 
estimating thin structure has received less attention.
This problem becomes more challenging in the presence of camera error, scene motion, and a constraint that curves are attached in a branching structure.
We  
propose a new general-purpose prior, a \emph{branching Gaussian processes} (BGP), that models spatial smoothness and temporal dynamics of curves while enforcing attachment between them.
We apply this prior to fit 3D trees directly to image data, using 
an efficient scheme for approximate inference based on expectation propagation.
The BGP prior's Gaussian form allows us to approximately marginalize over 3D trees with a given model structure, enabling principled comparison between tree models with varying complexity.  We test our approach on a novel multi-view dataset depicting plants with known 3D structures and topologies undergoing small nonrigid motion.  Our method outperforms a state-of-the-art 3D reconstruction method designed for non-moving thin structure.  We evaluate under several common measures, and we propose a new measure for reconstructions of branching multi-part 3D scenes under motion.

\keywords{multiview stereo, nonrigid models, expectation propagation}
\end{abstract}

%% file: intro.tex
\section{Introduction}

Curvilinear 3D tree structure is of central importance to several areas of science including the study of plants~\cite{Evers:2011uy,Trachsel:2011wu}, vascular systems~\cite{DenBuijs:2006jy,Nordsletten:2006fc,Strasser:2010ga}, and branching neurons~\cite{Cubelos:2010ic,Cuntz:2010dc}.  But estimating such 3D structure from images poses unique challenges, because the 3D shapes are thin, self-occluding, and often textureless. In addition, traditional reconstruction methods like visual hull or multi-view stereo require precise knowledge of camera poses and intrinsic parameters, and they assume the scene remains stationary between images.  When either of these are not true, epipolar constraints are violated and 3D reconstruction fails.  This can be especially problematic for multi-view 3D reconstruction of plants, because they can move a surprising amount in just a few minutes.  

We propose a model-based approach to recover 3D structure in these scenarios by explicitly modeling nonrigid motion and by defining a noise model that is robust to camera errors.
We introduce a new representation called a \emph{curve tree}, which defines both the curves' geometry and their topological structure. 
The curve tree representation provides more than just \emph{geometric} structure; the \emph{logical} structure is represented as well so that distinct parts may be identified, counted, and measured, and topology of the multi-part structure may be understood.
We then introduce a novel family of priors over curve trees called \emph{branching Gaussian processes} (BGPs) that model curves as smooth Gaussian processes and enforce attachment between curves in a tree topology.
We also extend the BGP to the temporal dimension to model 
sequences of curve trees over time.
This prior aids inference by limiting the space of plausible models, and in the case of a moving scene, it is necessary to make the problem well-posed.

Using the temporal BGP prior, we pose 3D reconstruction from images as Bayesian inference. Our image-based likelihood function is non-convex and highly multimodal, making exact inference intractable. We develop an approximate inference scheme based on expectation propagation that exploits the BGP prior's Gaussian form to approximate the posterior as multivariate Gaussian.
We also approximately solve the intractable Bayesian model selection problem, allowing us to compare curve trees of varying complexity in a principled way.
Our method is fully trained from a single exemplar sequence and is effective at recovering 3D branching structure even in scenarios with ambiguous data and self-occlusion. When ambiguity cannot be resolved, our method provides a rich representation of uncertainty in the form of a posterior distribution.  We evaluate our results using the DIADEM metric for branching structure and a new score designed to assess the logical structure of multi-part tree-structured 3D models.


%% file: related_work.tex
\section{Related Work}

%
Gaussian processes (GP's) are an increasingly common tool for machine learning and computer vision~\cite{Rasmussen:2006vz}, especially as smoothing priors for deformable geometry.
Zhu~\etal used GP's for recovery of nonrigid 3D surfaces~\cite{Zhu:2009ud}, and Serradell~\etal used GP priors to match graphs embedded in 2D or 3D~\cite{Serradell:2012kr}. 
Like us, Sugiyama~\etal proposed constructing a GP over an embedded graph~\cite{Sugiyama:2008ea}, but their covariance function is based on geodesic distance, which is not positive definite in general.
Our proposed covariance function is a positive-definite alternative but still has the property that correlations are inversely monotonic with geodesic distance.

Shape estimation from images has been widely studied, but less attention has been paid to the challenges posed by thin geometry.  
Segmentation of thin structure in images has gained some attention in recent years~\cite{Gu:2015in,Marin:2015vk,Tu:2006ga,Vicente:2008fe}, but the problem of 3D reconstruction remains largely unexplored.
%
Lopez~\etal used space carving to recover thin branching structure as a voxel volume, but their approach relies on near-exact camera calibrations~\cite{Lopez:2010tn}.
Amy Tabb achieved robustness to camera errors using a probabilistic variation of visual hull, allowing the position of thin structure to be inconsistent in some cameras and still be reconstructed~\cite{Tabb:2013da}.
Both of these approaches assume the scene is static between images, and they produce only geometry, not part labels or topology.
Kahl and August propose a generative model for 3D space curves that follow a Markovian random process through space~\cite{Kahl:2003et}.  They suggest their model could be made robust to both camera errors and scene motion by modeling small perturbations to the 3D curves in each image.  We achieve robustness in a similar way, but our model (a) permits a large class of Gaussian processes beyond just Markov processes, (b) introduces branching attachment between curves, (c) enforces temporal consistency between per-view perturbations, and (d) addresses the model selection problem when fitting an unknown number of curves. 

Our approach seeks not just 3D shape (\ie, geometry), but 3D \emph{structure} (\eg, topology,  medial axis, distinct geometric parts). 
Branching topology is of particular importance since it enables measurement of morphological traits such as branch angles,  branching depth, and average curvature of individual branches.
Approaches exist for estimating 2D trees from images~\cite{Turetken:2013in,Turetken:2012ef} and 3D trees from 3D data modalities like point clouds~\cite{Song:hi} and voxels (\eg, the Diadem challenge,~\cite{Bas:2011ib,Cho:2011,Turetken:2012ef,Turetken:2011fp,Wang:2011,Zhao:2011bv}).  Methods estimating 3D tree structure from 2D projective images are less common.
Some image-based methods recover a plausible branching 3D plant model without attempting to recover the exact branching architecture~\cite{Chen:2008kf,Diener:2006us,Li:2011kb,Talton:2011gd,Tan:2008dq,Tan:2007hf,Shlyakhter:2001tc}. In addition, several semi-automatic tools use human input to help reconstruct vascular systems~\cite{vmtk}, root system architectures~\cite{Clark01062011,LeBot:2009dv,zhu:2006}, and neuron arbors~\cite{Narro:2007hw} from images.

%
The motion of 3D trees has been modeled in computer graphics for generating animated synthetic trees (\eg,~\cite{Diener:2006us,Li:2011kb}).
 Estimating the motion of real trees is less common.  Glowacki~\etal recover sequences of 2D trees in a two-step process, using bottom-up methods to find a graph of tree points and integer programming to find the optimal tree-structured subgraph~\cite{Glowacki:2014gf}. 
However, their bottom-up approach does not address self-occlusions, and they rely on the graph deformation model of Serradell~\etal~\cite{Serradell:2012kr} based on Euclidean embeddings, which cannot model 2D curves that pass-by each other due to parallax.

While previous work addresses parts of our problem, we are unaware of other systems that attempt accurate estimation of 3D trees under nonrigid motion.  


%% file: bgp.tex
\section{Curve Trees}

We define a \emph{curve tree} as a finite set of curves attached in a tree topology.  The \(c\)\th curve in a \(d\)-dimensional curve tree is defined by the \emph{curve function} \(f_c : [0, L_c] \rightarrow \R^d\).  In the expression \(f_c(t)\), the value \(c \in \{1,\dots, N\} \) is the \textit{curve index}, and \(t \in [0,L_c]\) is the real-valued \textit{spatial index}.  The point \(f_c(0)\) is the curve's \textit{initial point}, and \(f_c(L_c)\) is its \textit{terminal point}.  
The Cartesian product of the curve index and spatial index is called a \textit{point index} \((c,t)\), and it uniquely identifies each point in a curve tree.  The set of all point indices is the \textit{index space} \(\I = \cup_{c=1}^N (\{c\} \times [0,L_c]) \cup (0,0)\), where \( (0,0)\) represents ``no point,'' which is included for notationally convenience.
    The collection of curve functions are combined into a \textit{curve-tree function} \(f : \I \rightarrow \R^d\), where \(f(c,t) = f_c(t)\) is the point on the \(c\)\th curve with spatial index \(t\).

If a curve's initial point lies on another curve, we say the curves are attached; the former curve is called a \textit{child curve}, and the point where it attaches on the \emph{parent curve} is called its \textit{branch point}.
Curves may have at most one parent curve, and curves with no parent are \textit{root curves}. 
Attachment between curves is represented by a \textit{branch function}.
The function \(\parentC : \{0,\dots,N\} \rightarrow \{0,\dots,N\} \) maps a curve index to the index of its parent curve or zero for root curves.  Similarly, \(\parentT : \{0,\dots,N\} \rightarrow \R \) maps the curve to the spatial index of its branch point on the parent curve, or zero for root curves. The \textit{branch function} combines \(\parentC\) and \(\parentT\): 
\begin{equation}
    \parent(c,t) = \begin{cases}
    (\parentC(c), \parentT(c)) & \text{ if } c \neq 0 \text{ and } \parentC(c) \neq 0  \\
    (0,0) & \text{ otherwise.}
    \end{cases}
    \label{eq:branch_function}
\end{equation}
The input \(t\) is unused, but including it simplifies the recursive definition later.
The parent/child attachment is stated formally by the attachment constraint: 
\begin{equation}
f(\xc,0) = f(\parent(\xc, t)), \label{eq:attachment_constraint}
\end{equation}
for all non-root curves \(\{c : \parent(x) \neq (0,0) \}\) and spatial indices \(t \in [0, L_c]\).  

\begin{figure}[t]
\centering
\includegraphics[width=\textwidth]{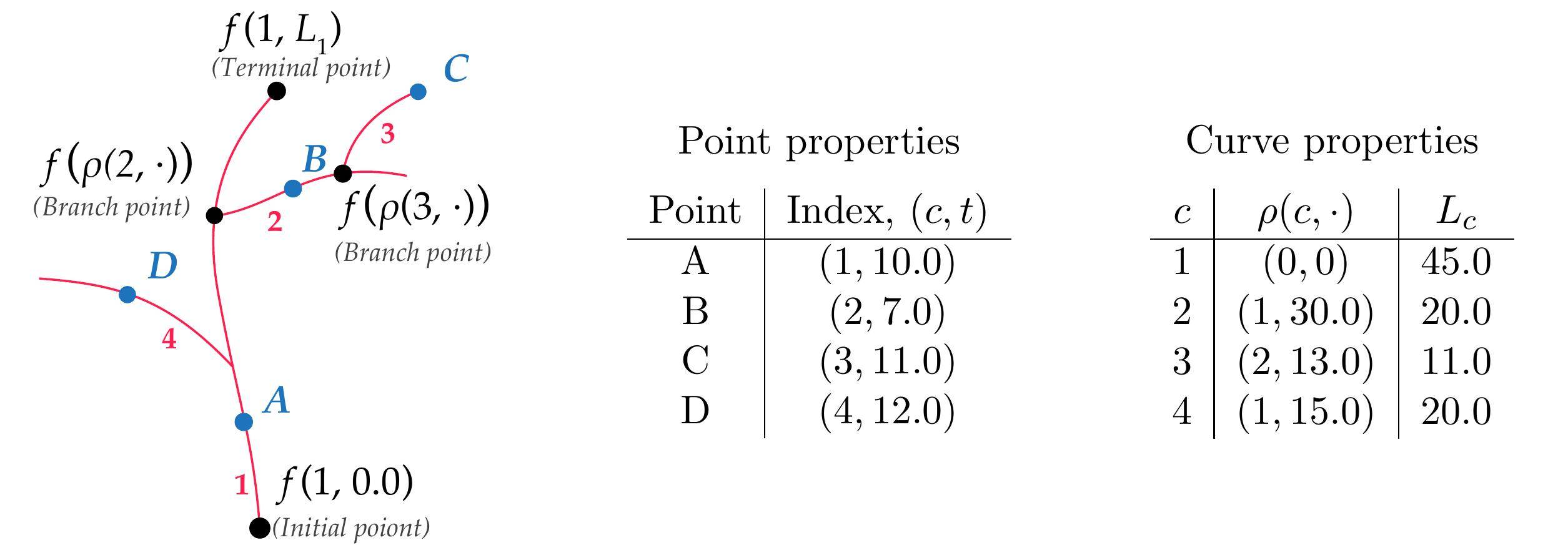}
\caption[Curve tree anatomy]{Anatomy of a curve tree (best viewed in color).  Curve indices are indicated in red. Initial, terminal, and branch points are noted.} 
\label{fig:curve_tree}
\end{figure}

A curve tree is fully defined by the number of curves \(N\), the spatial-input upper-bounds \(\bm{L} = \{L_1, \dots, L_N\}\), the curve-tree function \(f(c,t)\), and the branch function \(\parent(c,t)\).  \Cref{fig:curve_tree} illustrates these concepts.

\section{Branching Gaussian Processes}
\label{sec:bgp}
A Gaussian process is a generalization of the Gaussian distribution to parameterized collections of random variables.  These collections may be uncountably infinite, allowing a Gaussian process to define a probability measure over the space of continuous functions.  Gaussian processes are defined by the property that any finite subset of random variables has a multivariate Gaussian distribution~\cite{Rasmussen:2006vz}.
A Gaussian process \(f : \Theta \rightarrow \R\) is denoted by
\begin{equation}
f(\theta) \distras \GP(\mu(\theta), k(\theta, \theta')),
\end{equation}
where \(\mu(\theta)\) is the mean function, and \(k(\theta, \theta')\) is the covariance function.  The choice of covariance function dictates the properties of the Gaussian process such as smoothness, periodicity, or stationarity. Without loss of generality, we assume that \(\mu(\theta)\) is zero, so \(k(\theta, \theta')\) fully defines the Gaussian process.

    We propose a novel Gaussian process over curve trees called a \textit{branching Gaussian process} (BGP) by defining a covariance function over the curve tree input space \(\I\).    A BGP is defined by three properties: (a) individual curves are continuous; (b) the curve tree satisfies the attachment constraint~\cref{eq:attachment_constraint}; and (c) each subtree is conditionally independent of the rest of the tree given the subtree's branch point.  
Below we define a family of branching Gaussian process kernels called a \emph{rooted recursive kernel}, in which (a) a point's marginal variance is determined by its distance from a branch point, and (b) a child curve inherits the variance of its branch point.
This is a good model for a plant, where the root point is fixed and leaf points vary more than branch points.

It is also possible to define a family of BGPs in which curves are modeled by stationary covariance functions (\eg, the squared exponential) and the marginal variance of all points are equal (see Supplementary Material).

 \subsection{Rooted Recursive Kernels}
\label{sec:rrbgp}

We define a random function \(f : \R \rightarrow \R^n\) as \textit{rooted} \iff \(f(0) = \mathbf{0}\) almost surely (\ie, it is ``rooted'' to the origin).  A Gaussian process is rooted \iff its covariance function has the property \(k(0,t) = k(t,0) = 0\) for any \(t \in \R\).  Two examples of rooted covariance functions are the dot product covariance function, \(f(t,t') = t\,t'\), and the Weiner process covariance function, \(f(t,t') = \min(t,t')\).  An arbitrary covariance function \(k : \R^2 \rightarrow \R \) can be converted to a rooted covariance function \(k'\) by conditioning on the point at \(t=0\):
\begin{equation}
k'(t,t') = k(t,t') - k(t,0)\, k(0,0)^{-1} k(0,t')\,.\label{eq:stationary_to_rooted_xfm}
\end{equation}

Let \((c,t)\) and \( (c',t')\) be the indices of two points in a curve tree, and without loss of generality, assume the tree depth of \( (c,t)\) is greater than or equal to \( (c',t') \).  The \emph{rooted recursive} BGP kernel \(k_r : \I \times \I \rightarrow \R\) is defined as
\begin{equation}
k_r(c,t,c',t') = \begin{cases}
        0 & \text{ if } c = 0 \text{ or } c' = 0 \\
        \delta_{\xc\,\xc'} k_c(\xt, \xt') + k_r(\parent(c,t), c',t') & \text{ otherwise} 
    \end{cases}
\label{eq:recursive_cov}
\end{equation}
where \(\delta_{\xc\,\xc'}\) is the Kronecker delta.  The \emph{curve covariance} \(k_c : \R \times \R \rightarrow \R\) is any rooted covariance function, and it
defines smoothness properties of individual curves.
The Kronecker delta ensures the first term contributes zero covariance between different curves.  Inter-curve correlation comes from the second term, which is a recursion that replaces the first input with its branch index; if the resulting index is shallower than the second input, inputs are swapped so the first input is always at least as deep as the second.  This recursive term contributes a constant offset equal to the variance of the nearest common ancestor point.  Because \(k_c\) is rooted, it contributes no variance to initial curve points, so \(k(c,0,c,0) = k(\rho(c),c,0) = k(\rho(c), \rho(c))\), which implies the attachment property, \(f(c,0) = f(\rho(c,0))\).  

\subsection{Temporal Branching Gaussian Processes}

We may extend the curve tree model by introducing a temporal input \(\tau\) to the point index, \( (c, t, \tau) \in \I \times \R \). We construct a temporal BGP covariance function from the tensor product of a standard BGP covariance function and a temporal covariance function:
\begin{equation}
k_\tbgp(c,t,\tau,c', t', \tau') = k_\T(\tau, \tau') k_\bgp(c,t,c',t') \label{eq:tbgp_covariance}~.
\end{equation}
Here, \(k_\T\) is a covariance function over temporal inputs \(\tau, \tau' \in \R\), and \(k_\bgp\) is any BGP covariance function.

%% file: modeling.tex
\section{Application: Temporal 3D Tree Structure from Images}
\label{sec:application}

\begin{figure}[t]
\centering
\subfloat[\label{fig:arabidopsis}]{
\includegraphics[width=0.55\linewidth]{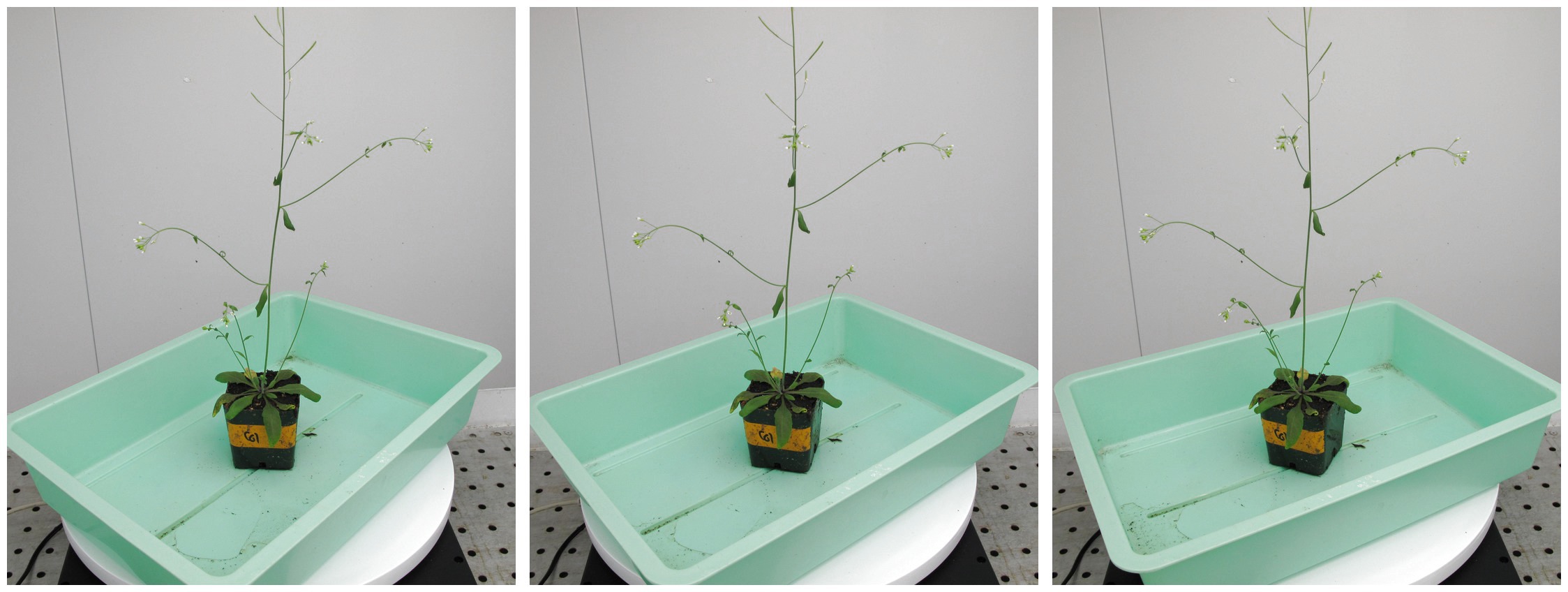}
}
\subfloat[\label{fig:plant_motion}]{
    \raisebox{0.75em}{
\includegraphics[width=0.2\textwidth]{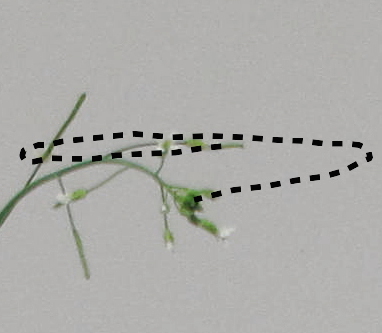}}
}
\subfloat[\label{fig:bad_cameras}]{
    \raisebox{0.25em}{
\includegraphics[width=0.2\textwidth]{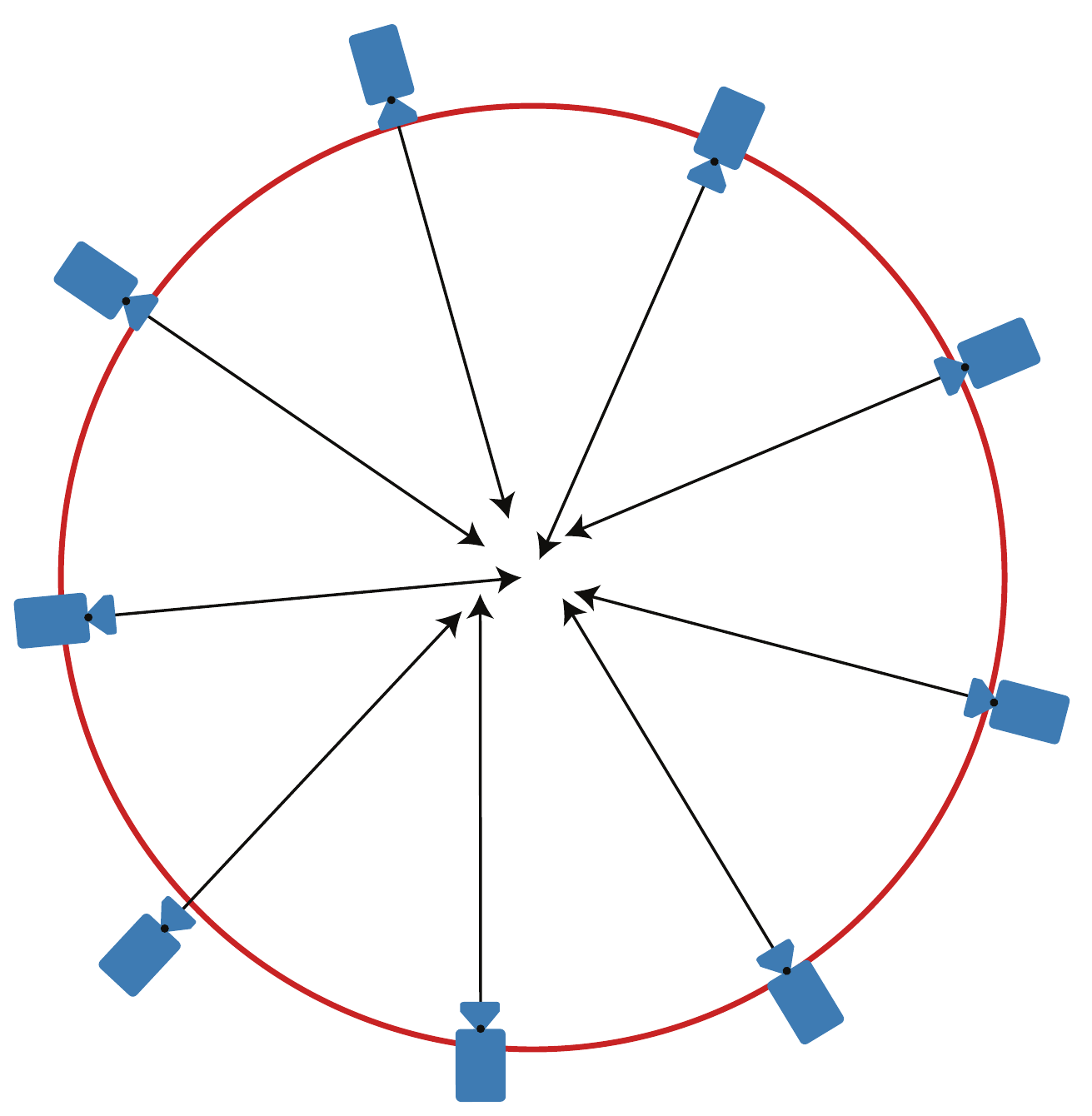}}
}
\caption{ \emph{(a)} First three images in an 18-view sequence, with ten degrees of rotation between views. \emph{(b)}  A stationary 3D point on a turntable should trace an elliptical arc in 2D; the non-elliptical path here indicates nonrigid motion. \emph{(c)} Turntable cameras should follow a circular path and have symmetric viewing directions.  This top-down illustration shows significant camera calibreation error, which complicates inference.  }
\label{fig:data}
\end{figure}

For the reconstruction task, we consider the first 18 images of an \Arabidopsis plant on a turnable separated by 10 degrees of yaw (\cref{fig:arabidopsis}) (the rear 18 images do not provide significant additional structural information). Images are captured at fixed time intervals, so time is represented by view-index. A camera model is estimated for each view using a crudely built calibration target, resulting in significant error (\cref{fig:bad_cameras}).   
Capture takes roughly three minutes, during which time the plant moves in response to changing direction of light and temperature gradients (\cref{fig:plant_motion}). Our goal is to recover a 3D curve tree for each of the 18 views.  
This problem is underconstrained, since variation in appearance between views may be explained by either parallax, nonrigid plant motion, or both.   
Our temporal BGP provides a prior over motion and geometry, making the problem well-posed.
We use a curve tree to represent the medial axis of a plant's stems.  Stems are mostly uniform in width, so we define the plant's shape as the union of spheres with 1\mm diameter and center \(f(c,t)\) for all \( (c,t) \in \I \setminus (0,0) \).  

\subsection{Prior}

The motion of plants tends to arise as random perturbations around a mean shape, and we define a BGP covariance function that reflects this:
\begin{equation}
k(c,t,\tau,c', t', \tau') =  k_\bgp(c,t,c',t') + k_\tbgp(c,t,\tau,c',t'\tau') + \sigma_0^2 . \label{eq:plant_covariance}
\end{equation}
The first term in \cref{eq:plant_covariance} is a branching GP covariance that represents a non-moving tree that is the center of perturbations.  We implement it using the rooted-recursive covariance function from \cref{eq:recursive_cov} with curve covariance
\begin{equation}
k_c(t,t') = \sigma_L^2 t\,t' + \sigma_s^2 \left( |t-t'| \frac{\min(t,t')^2}{2} + \frac{\min(t,t')^3}{3}\right)\,. \label{eq:plant_curve_covariance}
\end{equation}
This is the sum of a dot-product kernel plus a cubic-spline covariance function.  The cubic-spline process penalizes the curve's second derivative, 
     representing a bending energy that encourages straight lines.
     The cubic-spline covariance can only generate curves having zero initial slope, so we add a dot-product kernel to model initial slopes.

The second term in \cref{eq:plant_covariance} is the temporal BGP covariance from \cref{eq:tbgp_covariance} and it models random perturbations.  
For the \(k_\T\) term in \cref{eq:tbgp_covariance} we use the Ornstein Uhlenbeck covariance function,
\begin{equation}
k_\T(\tau,\tau') = \Exp{\abs{\tau-\tau'} / \ell_\T},  \label{eq:k_T}
\end{equation}
which is a mean-reverting random walk that ensures random perturbations never drift too far from the mean tree structure.  
    For the \(k_\bgp\) term in \cref{eq:tbgp_covariance} we again use a rooted recursive BGP covariance function with the same form as \cref{eq:plant_curve_covariance} but with different parameters \( (\sigma'_L)^2\) and \( (\sigma'_s)^2\).  

The final term in \cref{eq:plant_covariance} is a constant offset variance.  Because the first two terms are rooted recursive BGP covariance functions, the tree's root point has zero marginal variance under them.  Adding a constant offset variance allows trees to initiate from anywhere in \(\R^3\), not just at the origin.  
We model all three (x, y, z) dimensions as \iid under this prior.

Although the curve tree curves are continuous, in practice we perform inference on a discrete subset of points at indices \(X = \{(c_1, t_1), \dots, (c_M, t_M)\}\).  Let \(z_{ij} = f(c_j, t_j, \tau_i)\) be the \(j\)\th 3D point in the curve tree at time \(\tau_i\), and let the vector \(\bz_i = (z_{i1}^\top, \dots, z_{iM}^\top)^\top\) represent the discretely sampled curve tree at that time.  Let the vector \(\bZ = \{\bz_1, \dots, \bz_T\}\) represent the set of curve trees over all time indices.

Our choice of \(k_\T\) makes our model Markovian in the temporal dimension, so we can decompose the joint prior into
\begin{equation}
p(\bZ) = p(\bz_1) \prod_{i=2}^T p(\bz_i | \bz_{i-1}). \label{eq:prior}
\end{equation}
Since all factors are linear Gaussian, this represents a linear dynamical system (LDS),
\begin{equation}
    \bz_{i} = F_i \bz_{i-1} + \epsilon_i \text{~,} \label{eq:plants_lds}
\end{equation}
with transition matrix \(F_i\) and system noise \(\epsilon_i \distras \N(0, Q_i)\) defined as
\begin{align}
F_i &= \dddK_{*i}^\top \dddK_i^{-1} \text{~, } \\
Q_i &= \dddK_i - \dddK_{*i}^\top \dddK_i^{-1} \dddK_{*i} \text{~.} 
\end{align}
Here, \(\dddot{K}\) denotes the Kronecker product \(K \odot I_3\), and 
\begin{align}
(K_i)_{jk} &= k(c_j, t_j, \tau_i, c_k, t_k, \tau_i) \text{~,} \\
(K_{*i})_{jk} &= k(c_j, t_j, \tau_{i-1}, c_k, t_k, \tau_{i}) \text{~. }
\end{align}
The Kronecker product lifts the covariance matrix from modeling one output dimension to modeling three \iid dimensions.

We train the prior using maximum marginal likelihood~\cite{Rasmussen:2006vz}, using a gradient-free local optimizer (MATLAB's \texttt{patternsearch} function) and multiple initializations to avoid local maxima.  (See Supplementary Material for trained values.)

\subsection{Likelihood}
The likelihood function projects the plant's 3D shape into each view, rasterizes it into a binary foreground/background image, and evaluates each pixel against the input image.  
We trained a random forest classifier to identify foreground (\ie, stem) pixels using a single training image~\cite{hall2009weka}.  The classifier generates a map \(D\) whose \(i\)\th pixel has value \(d_{i} \in [0,255]\) representing the posterior probability of the class (foreground/background) given the evidence.  However, using posterior values in a likelihood function would introduce bias in favor of the more common class (\ie, background), so we next re-calibrate the classifier outputs to represent \emph{likelihood} values, \ie, the probability of the \emph{evidence}, given the \emph{class}.

Let \(\gamma^*_i \in \{0,1\} \) be value of the \(i\)\th binary pixel in the training foreground mask.
We construct a histogram over the set \(\{d_{i} | \gamma^*_{i} = 1\}\) that represents the likelihood of classifier output given a foreground label, \(p(d_i | \texttt{fg})\).  We construct a background likelihood \(p(d_i | \texttt{bg})\) similarly.  
To avoid overfitting, we smooth the histograms with a Gaussian filter with \(\sigma_{\mathrm{lik}}= 1\).   
The likelihood of the image at time \(\tau\) is then
\begin{equation}
p(D_\tau | \bz_\tau) = \prod_i p{(d_i | \mathtt{fg})}^{\gamma_{\tau i}} p{(d_i | \mathtt{bg})}^{1-\gamma_{\tau i}}. \label{eq:single_view_likelihood}
\end{equation}
where \(\gamma_{\tau i}\) is the value of the \(i\)\th binary pixel in the rasterization of \(\bz_\tau\) into view \(\tau\).
The supplementary materials describe an efficient method evaluating this likelihood.
The full likelihood over all images \(\D = \{D_1, \dots, D_T\}\) is the product of per-view likelihoods, \(p(\D | \bZ) = \prod_{\tau=1}^M p(D_\tau | \bz_\tau)\).

%% file: inference.tex
\section{Inference}

We begin by describing inference of a curve tree's 3D shape when its topological structure (\ie, the number of curves, point indices, and branch function) is known.  Later, we present methods for bootstrapping and refining the topology using Bayesian model selection.
The goal is to estimate a posterior distribution over points in the curve tree over all time frames.   Closed-form inference is not possible because the likelihood function is non-convex and highly multimodal.  However, the posterior distribution is likely to be highly peaked due to strong evidence from multiple views, making it well modeled by a multivariate Gaussian distribution.  Thus, we seek a Gaussian approximation of the true posterior that minimizes their \KL divergence.

\subsection{Estimating 3D Structure Given Topology}

Our inference approach assumes 3D points are initially observed by a single reference image, making them well-localized in two dimensions, but their depth is uncertain; sections \ref{sec:ep_bootstrapping} and \ref{sec:topology_refinement} describe methods for achieving such initializations.  
Under this assumption, inference is akin to an epipolar search in several views.
Conditioned on the reference view, a point's probability density is concentrated along a 3D ridge near the reference image's backprojection line. This ridge's width is a function of camera calibration error and scene motion, and its length is loosely constrained by the prior to be near the center of the scene.  We call the projection of this ridge the \emph{epipolar region} of a point in a particular view.
  The prior defines strong spatiotemporal correlations between points, so ambiguous evidence in a particular image can be resolved by consulting nearby points in space and time that are unambiguous.
This contextual information can shrink the ambiguous point's epipolar region, allowing some local optima to be ruled out, and making the point suitable for informing other ambiguous points.

We implement this logic using the expectation propagation algorithm (EP) where a Gaussian approximation of the posterior is iteratively updated, starting as noninformative and becoming more peaked as more evidence is considered~\cite{Minka:2013we}.  When the likelihood in an epipolar region has multiple modes, it is approximated by a Gaussian with high variance so all modes are well supported.  This effectively defers a decision on the point's location until later iterations when local context is more informative. 
%

Let \(f(\theta)\) be an arbitrary distribution and \(q(\theta)\) be a Gaussian approximation.  EP seeks parameters of \(q\) that minimize the Kullback-Leibler divergence:
\begin{equation}
    KL(f || q) = \int f(\theta) \log \frac{f(\theta)}{q(\theta)} d\theta \label{eq:ep_kl}.
\end{equation}
Our posterior distribution has the form
\begin{equation}
    p(\bZ | D) \propto p(\bZ) \prod_{i=1}^n L_i(\bZ),
\end{equation}
where \(p(\bZ)\) is the temporal BGP prior in~\cref{eq:prior} and \(L_i(\bZ) = p(D_i | \bz_i)\) are the single-view likelihood terms from \cref{eq:single_view_likelihood}.  
We seek a multivariate Gaussian approximation \(q(\bZ) \propto p(\bZ) \prod_{i=1}^T \hat{L}_i(\bZ)\), 
where \(\hat{L}_i(\bZ)\) are multivariate Gaussian approximations of \(L_i(\bZ)\) with diagonal covariance.

Standard expectation propagation involves estimating the first two moments of the joint posterior at each iteration.  In a curve-tree sequence with \(M\) points and \(T\) time frames, this would involve an intractable integral in \(3TM\) dimensions.  Instead, we adopt a variation of EP proposed by Gelman~\etal, where conditionally independent likelihood terms are updated independently in parallel in each iteration, rather than updating the entire joint posterior~\cite{Gelman:2014ww}.   If we assume the likelihood of each point is independent conditioned on the 3D model, Gelman's method reduces moment-estimation to \(TM\) integrals in three dimensions.  This independence assumption holds if we ignore self-occlusions which are minimal and take the limit as polyline point density approaches infinity.

Our EP implementation is shown in \Cref{alg:ep_updated}.
Here \(q_{ij}(z_{ij})\) is the single-point marginal of the approximate Gaussian posterior distribution, \(q(\bz_i)\).  We define the single-point likelihood \(L_{ij}(z_{ij})\) as the likelihood of a sphere centered at \(z_{ij}\) with diameter 1\mm, evaluated under the single-view likelihood in \cref{eq:single_view_likelihood}.  This single-point, single-view likelihood can be computed efficiently using a lookup table (see supplement).  
We approximate the moments of the three-dimensional marginal posterior of a point using importance sampling (again, see supplement).
Gelman's algorithm requires a global update after each pass, causing the joint posterior to reflect the updated likelihood approximations.  This would take \(O( M^3T^3)\) time if implemented na\"{\i}vely; instead, we exploit our model's LDS structure from \cref{eq:plants_lds} to implement the global update with a Kalman filter and smoother in \(O(M^3 T)\) time.  


\begin{algorithm}[t]
\caption{Our proposed variant of expectation propagation}
\label{alg:ep_updated}
\begin{enumerate}
\newcommand{\cav}[2]{#1^{-(#2)}}
\newcommand{\tlt}[2]{#1^{\setminus{(#2)}}}
\item Initialize \(\{\hat{L}_{ij}(z_{ij})\}_{i=1}^T\) to improper Gaussians for \(i \in \{1,\dots, T\}\) and \(j \in \{1,\dots,|\bz|\}\).

\item Initialize \(\{q_{i}(\bz_{i})\}_{i=1}^T\) to the marginal BGP priors, \(\{p(\bz_i)\}_{i=1}^T\).
\item For each \( (i,j) \in \{1,\dots, T\} \times \{1,\dots,|\bz|\}\): \label{st:sep_loop}
    \begin{enumerate}
        \item Compute the ``cavity'' distribution \(\cav{q_{ij}}{i,j}(z_{ij}) = q_{ij}(z_{ij}) / \hat{L}_{ij}(z_{ij})\). \label{st:sep_1}
        \item Compute the ``tilted'' distribution \(\tlt{q_{ij}}{i,j}(z_{ij}) = \cav{q_{ij}}{i,j}(z_{ij}) L_{ij}(z_{ij}) \). \label{st:sep_2}
        \item Update $\hat{L}_{ij}(z_{ij}) $ so \(\cav{q_{ij}}{i,j}(z_{ij}) \hat{L}_{ij}(z_{ij}) \) approximates \(\tlt{q_{ij}}{i,j}(z_{ij})\):  \label{st:sep_3} 
        \begin{enumerate}
            \item Construct a 3x3 Gaussian \(q_{ij}(z_{ij})\) using the first two moments of \(\tlt{q_{ij}}{i,j}(z_{ij})\). 
            \item Set \(\hat{L}_{ij}(z_{ij}) = q_{ij}(z_{ij}) / \cav{q_{ij}}{i,j}(z_{ij})\).
        \end{enumerate}
    \end{enumerate}
\item Use Kalman filter and smoother to re-compute the per-view marginal posteriors \(\{q_1, \dots, q_M\}\) using the new likelihood approximations \(\hat{L}_{ij}\). \label{st:sep_4}
\item Repeat step~\ref{st:sep_loop} until convergence.
\end{enumerate}
\end{algorithm}

\subsection{Bootstrapping}
\label{sec:ep_bootstrapping}

We bootstrap the inference process by using 2D image processing and backprojection to estimate an initial 3D tree whose depth is uncertain.
We first construct a binary image of likely foreground pixels in the first view consisting of pixels whose foreground likelihoods are higher than their background likelihoods.  The binary map is further refined by taking the median image over all views and removing any pixel whose intensity is not significantly different from the median pixel--\ie, basic background subtraction.  Next, we eliminate all but the largest connected component from the foreground image.  The result is a set of likely stem pixels with relatively few false positives.

We construct a 2D tree from this binary image by taking its morphological skeleton, constructing an adjacency graph of skeleton pixels, and extracting the graph's Euclidean minimum spanning tree.  Each point in the 2D tree is backprojected to a depth closest to the prior mean (\ie, the origin).  Chains of points with degree two or less are grouped into curves; points with higher degree become branch points.   Each curve's points are assigned spatial indices according to their arc-length parameterization in 3D.  In this way, we derive the topological structure of a 3D curve tree: the number of curves \(N\), curve lengths \(L_i\), point indices \(X\), and the branch function \(\rho(c,t)\).
This bootstrapping heuristic can miss some curves and mis-estimate branch points, but these errors may be fixed during the topology refinement process described in the next section.

We construct an initial likelihood approximation for the first view from the 2D tree points by
\begin{align}
    \hat{L}_{1j}(z_{1j}) &= \N(z_{1j}, \Lambda_{1j}^{-1}) \,~\text{ where} \label{eq:bootstrap_likelihood} \\
    \Lambda_{1j} &= H' (I_2/\sigma_L^2) H
\end{align}
Here, \(z_{1j}\) is the \(j\)\th backprojected 3D point, \(H\) is the Jacobian of the perspective projection of \(z_{1j}\) into the first view, and \(\sigma^2_L\) represents the uncertainty of the detected points.  We use \(\sigma^2 = 1 \text{ pixel}^2\), which represents error due to pixel quantization and small localization errors.  The precision matrix \(\Lambda\) is rank deficient, reflecting a lack of evidence for the point's depth.  

\subsection{Topology Refinement using Bayesian Model Selection}
\label{sec:topology_refinement}

\begin{algorithm}[t]
\caption{3D Tree Structure Estimation}
\label{alg:full_algorithm}
\begin{enumerate}
\item Initialize each \(\hat{L}_i(\bz_i), q_i(\bz_i)\) to degenerate Gaussians for \(i \in \{1, \dots, T\}\).
\item Bootstrap: estimate \(\hat{L}_1(\bz_1), q_1(\bz_1)\) using the procedure in~\cref{sec:ep_bootstrapping}.
\item Update \(\{q_2(\bz_2), \dots, q_T(\bz_T)\}\) with a Kalman filter pass.
\item Run three iterations of EP, keeping \(q_1\) and \(\hat{L}_1\) fixed on first pass. 
\item Run pruning.
\item For each time index {\(i \in \{5,9,13,\dots,T\}\)}:
\begin{enumerate}
    \item Run birthing with reference image \(i\).
    \item Run three iterations of EP.
    \item Run pruning.
\end{enumerate}
\end{enumerate}
\end{algorithm}


Trees with differing number of curves will have different dimensionality, so their posterior distributions are not comparable.  This is a model selection problem, and the Bayesian solution is to integrate out the continuous parameters (\ie, the curve points) and compare the models' marginal likelihoods.
The marginal likelihood does not over-fit and it includes a natural penalty for model complexity, making it an ideal model selection criterion.  Computing it involves a high-dimensional integral that is intractable, but we estimate it using the Gaussian approximation we constructed during expectation propagation.  

The marginal likelihood is \(p(\D|\mathcal{M})\), where \(\mathcal{M}\) represents curve tree's topological parameters, \ie, the number of curves \(N\), the point indices \(X\), and the branch function \(\rho(c,t)\).
  In what follows, we omit the dependence on \(\mathcal{M}\).
  Using the chain rule we rewrite the marginal likelihood as
\begin{equation}
p(\D) = p(D_1) \prod_{t=2}^T p(D_t | D_{1:i-1}). \label{eq:marg_lik_decomposed}
\end{equation}
Under Bayes' rule (after rearranging),
\begin{equation}
p(D_{i} | D_{1:i-1}) = p(D_i | \bz_i) \left( \nofrac{p(\bz_i | D_{1:i-1})}{p(\bz_i | D_{1:i})}  \right)\,. \label{eq:cond_lik}
\end{equation}

The first term in~(\ref{eq:cond_lik}) is the true single-view likelihood, \(L_i(\bz_i)\).  The second term is a ratio of the partial posteriors before and after considering data \(D_i\). Gaussian approximations of these distributions are computed during Kalman filtering before and after the update step, respectively.  
In theory, any value of \(\bz_i\) can be used to evaluate \cref{eq:cond_lik}, but our approximations are most accurate near the true posterior mode, so we use the current estimate of the mean of the approximate Gaussian posterior, \(q(\bZ)\).  

The space of tree topologies can be explored by proposing incremental changes to the current curve tree topology and keeping the model with the highest marginal likelihood.
We propose two such methods: pruning and birthing.

\textbf{Pruning.}
In the pruning step, a curve is deleted if doing so improves the marginal likelihood.  Each curve is considered for pruning in random order.  Pruning a parent curve causes its descendant subtrees to become independent trees.
Pruning a Y-junction results in two curves attached end-to-end.

\textbf{Birthing.} The birthing step mimics bootstrapping, but when constructing the binary mask, we omit foreground pixels 
if they coincide with the projection of existing structure.   
The remaining pixels are skeletonized, backprojected, and converted to 3D curve trees as in bootstrapping, and the root of each new tree is attached to the nearest point on existing structure.  Likelihood approximations are created for the new points in the reference view using \cref{eq:bootstrap_likelihood}.  We propagate this likelihood to the other views by a pass of Kalman filtering and smoothing.
A subsequent pruning step removes spurious structure added during birthing.

%
\Cref{alg:full_algorithm} shows how bootstrapping, geometry estimation, and topology refinement are applied in an end-to-end system.
In principle, expectation propagation should be run after each model modification before evaluating the marginal likelihood.  In practice, we omit EP after the pruning step, as we observed that it caused negligible change to the model and the approximated marginal likelihood.

%% file: experiments.tex
\section{Experiments and Results}

\begin{figure}[t]
\subfloat[]{
\includegraphics[width=0.23\linewidth]{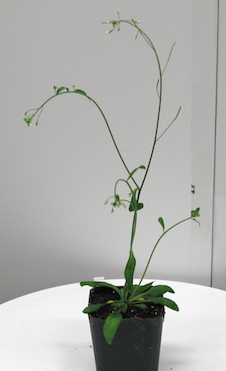}}
\subfloat[]{
\includegraphics[width=0.23\linewidth]{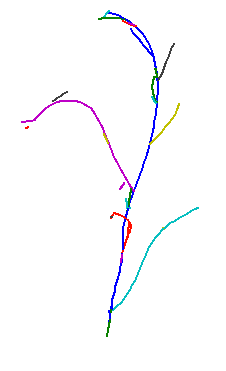}}
\subfloat[]{
\includegraphics[width=0.23\linewidth]{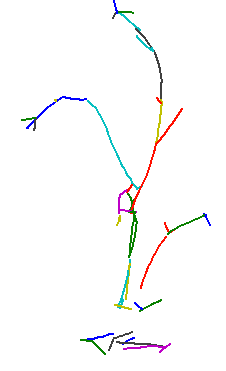}}
\subfloat[]{
\includegraphics[width=0.24\linewidth]{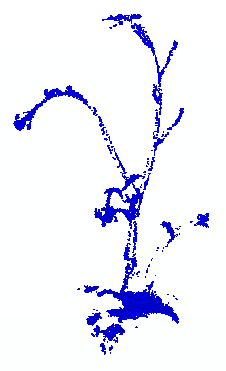}}
\caption{Results of 3D reconstruction.  \emph{(a)} One of 18 input images.  \emph{(b)} Semi-automatic reconstruction result.  Color indicates distinct parts. \emph{(c)} Fully-automatic reconstruction result. \emph{(d)} Voxel volume produced by~\cite{Tabb:2013da}.  
}
\label{fig:results}
\end{figure}

We applied our approach to recover 3D structure of plants from images.  
Our dataset consists of image sequences of twelve \Arabidopsis plants as described in \cref{sec:application}.  One specimen was held out for training, the rest were used for testing.  
Ground truth was manually collected on every fourth image by tracing each stem with a 2D \polybezier curve.
Curve identities were kept consistent across views, and parent/child relationships between curves were noted.  Each \polybezier curve was sampled at one pixel intervals
and each point in each image was backprojected to a depth that minimized the sum of squared distances between its 2D projection and the nearest ground truth curve point in each image.
This results in a sequence of five 3D curve trees per specimen that project exactly to their 2D counterparts.  

We evaluate using four measures.  We compute a one-to-one matching between points with distance at most 1\cm (see supplement). We report accuracy as the RMS error of matching points and we measure completeness and spurious structure using an intersection-over-union score that compares the number of matched points against the total number of points in both models.   We evaluate tree topology using the DIADEM metric with a threshold of 1\cm in the x, y, and z directions~\cite{Gillette:2011ha}.  However, the DIADEM metric does not distinguish ``parts,'' so it cannot distinguish between Y- and T- junctions, and it does not recognize part-level structural errors like over-segmentations of curves or misclassified T-junctions.  Further, it has no concept of time, so identity switches are not penalized.  

To address these issues, we propose a \emph{tree-structure consistency} (TSC) score that measures structural consistency in three ways: (a) the geometric consistency of matching curves, (b) matching consistency between child and parent curves, and (c) consistency of curve identity between adjacent time frames.  Evaluation begins by finding a correspondence between parts that maximizes their number of matching points. Then consistency is measured with respect to the corresponding parts.  The TSC score measures the number of logical errors in the structure, normalized to the range \([0,1]\), where \(1\) represents ``no errors.''  For details, see the supplementary material.

We ran our algorithm twice, once using all 18 images, and once using only the first nine.  We also compared two variants of bootstrapping: a fully-automatic approach as described in \cref{sec:ep_bootstrapping}, and a semi-automatic approach where bootstrapping is replaced with a manually traced 2D curve tree in the first image.    
Inference took an average of 33 minutes using Matlab code on a 16-core Xeon E5-2650 CPU running at 2.6GHz.
Results for all runs are shown in \Cref{tab:results}.  
We see that semiautomatic reconstruction performs significantly better on the TSC score than the fully automatic one, primarily due to the latter oversegmenting curves.  This could be addressed with a ``join curves'' step during topology refinement.
\Cref{fig:results} provides a visual comparison against Tabb's Shape-from-Silhouette-Probability-Maps (SfSPM) algorithm, which produces volumetric reconstructions of thin structure~\cite{Tabb:2013da}.  (For visualizations of all eleven reconstructions, see supplementary material.)


\begin{table}[t]
\caption[Reconstruction results]{Structure estimation performance of semi- and fully-automatic reconstruction, for nine- and 18-image datatasets. }
\scriptsize
\begin{center}
\setlength{\tabcolsep}{2.8mm}
\begin{tabular}{r | c  c  c c  }
\hline
& Acc & IoU & TSC & DIADEM~\cite{Gillette:2011ha} \\
\hline
\hline
 Semi (9) & \(4.7 \pm 1.4 \) & \( 0.63 \pm 0.14 \) & \( 0.49 \pm 0.09 \) & \( 0.81 \pm 0.09 \)\\
Semi (18) & \(3.7 \pm 1.0 \) & \( 0.70 \pm 0.08 \) & \( 0.56 \pm 0.06 \) & \( 0.84 \pm 0.06\)\\
\hline
Auto (9) & \(7.1 \pm 3.1 \) & \( 0.50 \pm 0.11 \) & \( 0.41 \pm 0.07 \) & \( 0.41 \pm 0.18 \)\\
Auto (18) & \(5.9 \pm 2.2 \) & \( 0.59 \pm 0.11 \) & \( 0.43 \pm 0.08 \) &  \( 0.44 \pm 0.18 \)\\
\hline
\end{tabular}
\end{center}
\label{tab:results}
\end{table}

%% file: conclusion.tex
\section{Conclusion}
   
We have demonstrated a novel probabilistic model for branching multidimensional structure, and a Bayesian method for 3D reconstruction of moving trees from a single moving camera.  
Our approach obtains good results with limited training data, showing that it generalizes well and avoids overfitting.
Our model selection scheme naturally penalizes complexity with no additional tunable parameters.   
By taking advantage of the model structure, our expectation propagation scheme locates strong modes reasonably efficiently and should outperform black-box Bayesian inference like hybrid Monte Carlo.
Our proposed BGP covariance function is a general-purpose kernel and can be applied in kernel methods such as SVM, PCA, GP classification, and kernel density estimation.
It is also straightforward to extend branching curves to branching surfaces;
this could be used to model deformable leaves attached to the curve tree.
We also contributed a new score measuring logical consistency of multi-part branching structure.  
Our dataset and evaluation code may be downloaded from ivilab.org.

 \textbf{Acknowledgments:}
 We would like to thank Dr. Amy Tabb for sharing her Shape from Silhouette Probability Maps code.  This material is based upon work supported by the National Science Foundation under Award Numbers DBI-0735191 and DBI-1265383 and the Department of Education’s GAANN Fellowship through the University of Arizona's Computer Science
  Department.
%
